%% file: iclr2025_conference.tex
\documentclass{article} 
\usepackage{iclr2025_conference,times}

\input{math_commands.tex}

\usepackage{hyperref}
\usepackage{url}
\usepackage{amsmath} 
\usepackage[utf8]{inputenc} 
\usepackage[T1]{fontenc}    
\usepackage{booktabs}       
\usepackage{amsfonts}       
\usepackage{nicefrac}       

\usepackage{microtype}      
\usepackage[table]{xcolor}
\definecolor{darkgrey}{rgb}{0.4,0.4,0.4} 
\usepackage{graphicx}

\newcommand{\hv}[0]{\ensuremath{\boldsymbol{h}} }

\newcommand{\xv}[0]{\ensuremath{\boldsymbol{x}} }

\newcommand{\zv}[0]{\ensuremath{\boldsymbol{z}} }
\newcommand{\Av}[0]{\ensuremath{\boldsymbol{A}} }
\newcommand{\Bv}[0]{\ensuremath{\boldsymbol{B}} }

\newcommand{\Wv}[0]{\ensuremath{\boldsymbol{W}} }
\newcommand{\Xv}[0]{\ensuremath{\boldsymbol{X}} }

\newcommand{\muv}[0]{\ensuremath{\boldsymbol{\mu}} }

\newcommand{\sigmav}[0]{\ensuremath{\boldsymbol{\sigma}} }

\usepackage{enumitem}
\usepackage{multirow} 
\usepackage{caption}
\usepackage{tabularx}
\usepackage[table]{xcolor}
\usepackage{makecell}  

\title{Semantic-guided LoRA Parameters Generation}

\author{Miaoge Li$^{1}$, Yang Chen$^{1}$, Zhijie Rao$^{1}$, Can Jiang$^{2}$, Jingcai Guo $^{1}$\thanks{ Jingcai Guo is the corresponding author.} \\
$^1$Department of Computing/LSGI, The Hong Kong Polytechnic University, Hong Kong SAR\\
$^2$School of Artificial Intelligence, Xidian University, China\\
\texttt{jc-jingcai.guo@polyu.edu.hk} \\
}

\usepackage{hyperref}    
\iclrfinalcopy 
\begin{document}

\maketitle

\begin{abstract}
Low-Rank Adaptation (LoRA) has demonstrated strong generalization capabilities across a variety of tasks for efficiently fine-tuning AI models, especially on resource-constrained edges. However, in real-world applications, edge users often exhibit task-specific preferences that are difficult to handle with a unified model trained under a closed-world assumption, and the challenge may further increase when there are significant domain shifts between training and deployment. Meanwhile, retraining/fine-tuning models for each user is also impractical due to its cost-intensive nature and privacy concerns over raw data utilization from edges. To address these challenges, we propose \textbf{Semantic-guided LoRA Parameter Generation (\texttt{\texttt{SG-LoRA}})}, the first of its kind framework to efficiently produce user-specific LoRA parameters without any additional training on user tasks or access to user-specific data. 
Concretely, \texttt{SG-LoRA} uses task descriptions as the semantic bridge, measuring their proximity to a set of known expert tasks in a shared embedding space. Based on this semantic guidance, it models the target task's LoRA parameter distribution to generate high-performing parameters for novel tasks. 
\texttt{SG-LoRA} enables the real-time construction of LoRA models aligned with individual intents by distilling knowledge from prominent LoRA experts and, meanwhile, offering a privacy-preserving solution for personalized model adaptation in a novel zero-shot open-world setting proposed in this work. 
%
Extensive experiments on multiple challenging tasks confirm the superior performance and remarkable adaptability of \texttt{SG-LoRA}. Code is available at \href{https://github.com/keepgoingjkg/SG-LoRA}{https://github.com/keepgoingjkg/SG-LoRA.}

\end{abstract}

\section{Introduction}

In recent years, deep learning has seen remarkable progress, largely driven by the advent of large-scale pre-trained models (LPMs) \cite{chung2024scaling,li2022blip,liu2023visual,rombach2022high,touvron2023llama}. Trained on massive and diverse datasets, these models demonstrate exceptional performance across a wide range of downstream tasks \cite{shenaj2024lora,alayrac2022flamingo,touvron2023llama}. However, as both model and data scales continue to grow, retraining the entire model becomes increasingly computationally expensive and often infeasible in practice. To mitigate this challenge, parameter-efficient fine-tuning (PEFT) methods have garnered considerable attention \cite{zhang2023llama,zhang2023adalora,ding2023parameter}. Among them, Low-Rank Adaptation (LoRA) \cite{hu2022lora} has emerged as a prominent approach. LoRA adapts pre-trained models by introducing a small number of trainable low-rank matrices into existing layers, achieving strong task-specific performance while leaving the original model weights unchanged.

While an increasing number of pre-trained LoRA modules are becoming publicly available, effectively leveraging them in real-world scenarios remains a significant challenge. The abundance of LoRA resources does not inherently guarantee robust generalization—particularly when the target domain diverges substantially from those seen during training \cite{qorbani2025semantic}. Simultaneously, user-side demands are becoming increasingly dynamic and personalized, particularly in edge environments where data privacy constraints and limited computational resources make large-scale retraining infeasible. These trends raise two critical questions:
{\textit{(1) How can LoRA-enhanced models be rapidly and efficiently adapted for on-device deployment?
(2) How can such adaptations accurately reflect individual user intent without requiring extensive retraining?}}


\begin{figure}[!t]
\centering
\includegraphics[width=1.0\linewidth]{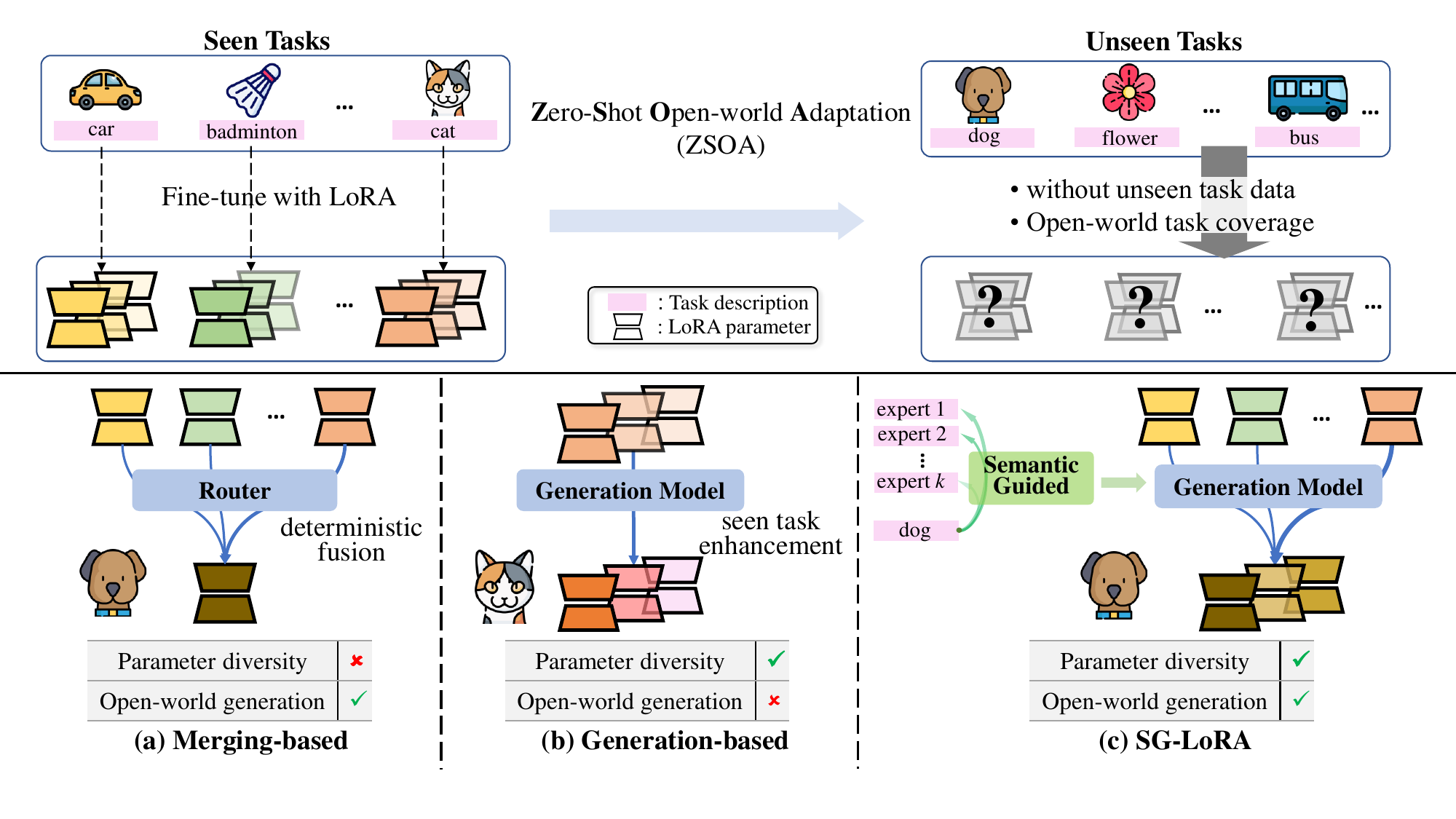}
\caption{\small{\textbf{Motivation of our \texttt{SG-LoRA}.} We consider a challenging scenario termed Zero-Shot Open-World Adaptation (ZSOA), where a model is provided rich LoRA resources for seen tasks but lacks access to data for unseen tasks during inference, with an unconstrained task space. Conventional LoRA adaptation methods are not suitable for ZSOA: merging-based approaches struggle to explore the diversity of LoRA parameters, while generation-based methods primarily focus on LoRA enhancement for seen task. Our \texttt{SG-LoRA} uses task descriptions as semantic guidance to enable conditional LoRA generation for unseen tasks in a data-free and open-world manner. (Each color family represents a set of LoRA parameters for the same task, for example, brown represents LoRAs for the \textit{'Dog'} task, and yellow represents LoRAs for the \textit{'Car'} task.)}}
\label{motivation_pic}
\end{figure}

To tackle these challenges, as shown in Figure.\ref{motivation_pic}, prior research has explored two main directions in the broader application of LoRA, each targeting one of the issues.
The first line of work focuses on fusion-based methods, which aim to rapidly construct task-specific models by directly merging existing LoRA modules at hand \cite{shenaj2024lora}. This approach eliminates the need for additional fine-tuning, enabling fast adaptation. However, such deterministic fusion is inherently limited in flexibility and generalizability: once the merged weights are determined, the resulting model is fixed to a specific task, restricting its ability to adapt to diverse or evolving requirements. Moreover, the merging process must be carefully designed, as conflicts may arise when integrating LoRA modules trained on different tasks \cite{zou2025cached,zhao2024merging}.
In parallel, another direction explores generation-based methods, which leverage generative models—such as variational autoencoders (VAEs) \cite{kingma2013auto} or diffusion models \cite{ho2020denoising}—to synthesize new LoRA parameters. By introducing stochasticity, these approaches enable greater diversity in parameter generation and bypass traditional fine-tuning pipelines. 
However, their success often relies on a closed-world assumption, where training and test tasks are drawn from similar distributions \cite{soro2024diffusion}.
As a result, these methods are susceptible to task (or domain) shifts—scenarios in which the data distribution at test time differs from that observed during training—often resulting in significant performance degradation. 

To address both challenges simultaneously, we introduce Zero-Shot Open-world Adaptation (ZSOA), a novel evaluation setting that assesses a model’s ability to generalize to previously unseen 
heterogeneous tasks, which share the same structural format(e.g., classification or retrieval) but differ in data distributions, without requiring access to labeled data from the target task.
ZSOA emphasizes two key aspects: (1) Open-world task coverage, where user-defined tasks emerge from a broad and unconstrained task space, maintaining structural consistency while varying in content, domain, or distribution.
(2) Adaptability to evolving user intents, reflecting real-world scenarios in which user needs change dynamically.

In this work, we draw inspiration from the human ability to intuitively infer semantic relationships between prior knowledge and new tasks \cite{lake2017building}, enabling effective generalization in unfamiliar situations. For example, after learning to recognize cat breeds such as \textit{Birman} and \textit{Egyptian Mau}, a person can identify \textit{British Shorthair} based solely on its textual description by relating it to previously acquired concepts. Motivated by this analogy-driven reasoning process, we propose a \textbf{S}emantic-\textbf{G}uided \textbf{LoRA} Parameter generation framework (\texttt{SG-LoRA}) that adapt LPMs to the ZSOA setting. Specifically, we begin by aggregating LoRA modules fine-tuned on known task-specific datasets to establish a repository of task experts. Rather than fine-tuning new LoRA parameters for each novel task, our method exploits task relationships embedded in their textual descriptions. These descriptions, typically concise, yet semantically rich, are processed by a frozen CLIP text encoder to capture task-level correlations without exposing user-specific data. More importantly,  simply scaling the number of expert LoRAs does not guarantee performance gains, as they may provide contradictory or irrelevant task knowledge. To resolve this, we design a sparse router that selectively combines the most semantically relevant expert LoRAs for a target task, thereby integrating rational prior knowledge for generating new task-adaptive LoRA parameters. We then train a generative model that leverages this acquired prior LoRA knowledge as conditional input to synthesize high-performing task-specific parameters. During inference, the system directly generates target LoRA modules aligned with user requirements using only prior knowledge. Crucially, the stochastic nature of our trained generator transforms deterministic LoRA construction into probabilistic parameter sampling—enhancing coverage diversity to dynamically adapt to evolving user intents. In summary, the main contributions of this work include 
:
\begin{itemize}
    \item {We introduce Semantic-Guided LoRA Generation, a versatile framework that harnesses semantic task relationships to enable zero-shot open-world adaptation. \texttt{SG-LoRA} conditions on prior task knowledge to synthesize high-performance LoRA parameters for arbitrary unseen tasks without retraining. 
    }
    \item By seamlessly integrating efficiently generated LoRA modules into off-the-shelf LPMs, our method enables fast personalization at inference time. It supports flexible updates to the expert database and enables exploration of diverse LoRA configurations for broad task coverage. Additionally, the framework is inherently transparent and controllable, providing a scalable solution for personalized and adaptable model behavior.
    \item Comprehensive experiments on multiple image-text retrieval benchmarks demonstrate that the proposed method can rapidly generate LoRA parameters with performance comparable to that of traditional LoRA fine-tuning.
\end{itemize}

\section{Related Work}


\subsection{Low-Rank Adaptation}
Low-Rank Adaptation
(LoRA) is a parameter-efficient fine-tuning method to adapt large models to novel 
tasks and shows superior performance \cite{huang2023lorahub,zhang2023composing,sung2022vl}. By optimizing new sets of low-rank matrices to approximate the weight changes during fine-tuning, LoRA dramatically reduces trainable parameters—maintaining knowledge from pre-training while minimizing computational overhead. Specifically, LoRA observes that a pre-trained model’s weight matrix often has a low intrinsic rank. Given a pre-trained weight $\Wv_{0}$ in the network, with $\xv$ as the input and $\hv$ as the hidden state, LoRA decomposes the
weight update $\Delta\Wv \in \mathbb{R}^{m \times n}$ into two low-dimensional matrics $\Bv \in \mathbb{R}^{m \times r}$ and $\Av \in \mathbb{R}^{r \times n}$:
\begin{equation}
\hv = \Wv_0\xv + \Delta\Wv\xv = \Wv_0\xv + \alpha\Bv\Av\xv,
\label{lora}
\end{equation}
where the rank $r \ll \min(n,m)$. The scalar $\alpha$ is a constant scaling hyperparameter that controls the contribution of the LoRA update in subsequent layers. Recently, a growing number of high-quality, pre-trained LoRA modules have become publicly available for various architectures \cite{civitai2024,huggingface2024}, including transformers and vision–language models, providing rich resources to accelerate adaptation and deployment on specific downstream tasks.



\subsection{Model Merging}

Model merging integrates parameter-level knowledge from multiple independently trained neural networks into a unified model, achieving enhanced capabilities \cite{li2023deep,ilharco2022patching,jang2024model,wortsman2022robust}. 
The pioneer work Model Soups \cite{wortsman2022model} establishes weight averaging as a foundational paradigm, demonstrating that averaging fine-tuned models from identical pre-trained bases with varied hyperparameters consistently outperforms individual models. Its efficacy arises from geometric alignment within shared loss landscape basins, where fine-tuned parameters converge to flat minima, coupled with the near-orthogonality of task-specific adjustments (task vectors) that minimizes interference during fusion.
AdapterSoup \cite{chronopoulou2023adaptersoup} generalizes the Model Soups paradigm to cross-domain adaptation by dynamically averaging domain-specific adapters at test time. This approach preserves the base model’s integrity while enhancing out-of-distribution generalization through selective weight-space interpolation of relevant domain knowledge. Recent advances have extended this paradigm to LoRA-based module fusion. For instance, 
LoraHub \cite{huang2023lorahub} dynamically composes pre-trained LoRA modules by optimizing their weights through few-shot examples from new tasks, leveraging black-box optimization techniques (e.g., CMA-ES) to achieve efficient adaptation without backpropagation. 
Meanwhile, SemLA \cite{qorbani2025semantic} introduces a training-free approach by directly comparing test images' visual features with known domain prototypes, using this similarity measure as an efficient domain navigator for both adapter retrieval and fusion. However, both methods are limited—either needing unknown-task data or employing inflexible deterministic fusion for new tasks.

\subsection{Neural Network Parameters Generation}  
Despite significant advancements in generative modeling, recent work has only begun to address the direct generation of network weights for pre-trained models \cite{knyazev2021parameter,zhmoginov2022hypertransformer,knyazev2023can}. Approaches such as generative hyper-representation learning \cite{ha2016hypernetworks}, neural network diffusion \cite{peebles2022learning,hu2021p,jin2024conditional}, and kernel density estimation–based methods \cite{soro2024diffusion} have shown promise but remain fundamentally limited to small architectures and unconditional weight generation within fixed distributions. Consequently, 
these methods are hampered in their ability to generalize to unknown tasks, constraining their broader applicability.
While meta-learning frameworks \cite{nava2022meta,zhang2024metadiff} have enabled powerful joint model generation for visual recognition and few-shot learning, they typically neglect the personalization and diversity of the generated parameters and restrict the generator’s output to classifier heads rather than more flexible and expressive parameter sets, such as LoRA. \cite{shao2025context} ICM-LoRA innovatively explores the parameter relationships among tasks through task vectors, but focuses on closed-world, task-specific enhancements of LoRA parameters. As a result, the question of whether one can rapidly generate efficient, user-intent–focused LoRA parameters in open-world settings remains unexplored.


\section{The Proposed Model}
In this section, we begin with an overview of essential concepts for understanding semantic-guided LoRA parameter generation,
followed by detailed descriptions of our proposed methods.


\subsection{Problem Defination and Preliminary}


We define Zero-Shot Open-world Adaptation (ZSOA) as a novel and challenging task setting that requires models to generalize across 
semantically diverse tasks in open-world scenarios. Unlike conventional zero-shot learning or task transfer—often confined to fixed label spaces or narrow domains—ZSOA focuses on tasks that share a common structural format (e.g., image-to-text retrieval) but differ substantially in domain, content, or distribution. 
It emphasizes that queried tasks during inference time are drawn from an undefined and unbounded set, requiring rapid adaptation to novel tasks without access to labeled data. This setting reflects realistic deployment scenarios, where task-level generalization must rely solely on 
prior experience.

In this work, we instantiate ZSOA in the context of fine-grained image-text retrieval, where each task corresponds to a retrieval problem defined by a specific semantic category (e.g., bird species, flower type). Formally, given a set of fine-tuned LoRA module $\mathcal{W}$ trained on known tasks $\mathcal{T}$, our goal is to adapt the model to unknown task $\mathcal{T}^*$ without accessing any labeled image-text pairs. 
Let $f(\mathcal{T})$ denote the textual description of task $\mathcal{T}$, we learn a generator $G$ that predicts LoRA parameters for $\mathcal{T}^*$ based on semantic descriptions and
$\mathcal{W}$:
\begin{equation} \label{cross-atten}
\mathcal{W}^* = G\left(f(\mathcal{T}^*), \mathcal{W}, f(\mathcal{T}) \right) 
\end{equation}
The synthesized LoRA parameter $\mathcal{W}^*$ is then used to modulate a frozen vision-language backbone, enabling it to perform the image-text retrieval task defined by $\mathcal{T}^*$. ZSOA thus extends traditional zero-shot learning by enabling parameter-level generalization 
, 
accommodating diverse user-instructed tasks with open-world queries.

\subsection{LoRA Parameter Dataset Construction}
\subsubsection{Task-Specific LoRA Training}\label{ss:loratraining}
The first stage involves constructing a dataset of LoRA parameters. Consider a collection of $N$ distinct tasks $\mathcal{T} = \{ \mathcal{T}_1, \ldots, \mathcal{T}_N \}$, where each $\mathcal{T}_i$ corresponds to a specific image-to-text retrieval task. In this context, the task is dataset-agnostic—that is, it may originate from the same dataset or from a different one.
Given a pre-trained vision-language model (VLM), we train a task-specific LoRA for each $\mathcal{T}_i$ using corresponding image-caption pairs, applying LoRA modules at consistent positions within the VLM. To ensure comparability and reduce variance, all LoRA modules are trained using identical network configurations by default. After training stabilizes, we extract and store the LoRA parameters from the final $M$ epochs, yielding task-specific parameter data:
\begin{equation} \label{lora_data}
\Delta\mathbf{\Wv}_i = \{ \Delta\mathbf{\Wv}_i^j \}_{j=1}^M, \quad \mathbf{d}_i = f(\mathcal{T}_i),
\end{equation}
where   $\Delta\mathbf {\Wv}_i^j$ is the $j$-th epoch LoRA parameter for task 
$\mathcal{T}_i$. $\mathbf{d}_i$ denotes the textual description associated with
$\mathcal{T}_i$, generated using the template \textit{'a photo of a <class name>'}, and encoded by the frozen CLIP text encoder $f(\cdot)$ to serve as a global semantic representation. Collectively, these form the LoRA parameter dataset:
\begin{equation} \label{lora_dataset}
\mathcal{W} = \{ \Delta\Wv_1, \ldots, \Delta\Wv_N \},
\end{equation}
where element is optimized for image-text alignment within its respective task.


\subsubsection{LoRA Expert Repository Formation}
To construct a reliable expert LoRA space, we first curate a representative subset of tasks from the full corpus, where each task is associated with its corresponding LoRA parameters and semantic embedding. For each selected task, we compute the mean LoRA parameters $\muv_i$ across all available adaptations, forming a distilled representation of the task-specific adaptation pattern. These mean parameters, along with their associated semantic embeddings, constitute our expert repository:
\begin{equation}\label{expert_sim}
\mathcal{W}_{\text{expert}} = \{(\boldsymbol{\mu}_i, \mathbf{d}_i) \mid i \in \mathcal{I} \}, \quad  \boldsymbol{\mu}_i= \frac 1 M\Delta\mathbf{\Wv}_i,
\end{equation}
where $\mathcal{I}$ represents the selected expert index and $ \mathcal{W}_{\text{expert}}$ serves as a compact yet expressive basis for capturing the essential characteristics of each knowledge domain. The remaining LoRA data are then partitioned into training and evaluation sets to facilitate subsequent model development and evaluation. 


\subsection{Semantic-guided LoRA Parameter Generation}

Our \texttt{SG-LoRA} framework generates conditional LoRA parameters under semantic guidance. We first construct a semantic prior by selecting and combining the most relevant expert LoRAs from the repository using a sparse router. A conditional variational autoencoder (CVAE)\cite{sohn2015learning} is then trained to generate target LoRA parameters aligned with the task description intent. Once trained, the model can generalize to new tasks in an open-world query setting without requiring further training.


\subsubsection{Construction of Semantic Prior}
Intuitively, not all experts contribute equally to an unknown task. Therefore, it is essential to identify and prioritize the most beneficial experts. Fortunately, the CLIP textual encoder is well-suited for this purpose, as it effectively captures rich semantic relationships across diverse tasks. As established in Eq. \ref{lora_data}, we use the textual embedding of each task as a global semantic descriptor for its corresponding LoRA parameters. For an unseen task $\mathcal{T}^*$ with textual embedding $\mathbf{d}^*$, we calculate the cosine similarity between $\mathbf{d}^*$ and the textual embedding $\mathbf{d}_i$ of each expert task.
We then select the top-$k$ experts with the highest similarity scores to form a semantically tailored expert set, indexed by $\mathcal{I}_{\text{top-}k}$. The LoRA parameters associated with these experts are fused to construct a prior for the unseen task $\mathcal{T}^*$.
Then, we normalize their similarity scores using the softmax function to obtain the fusion coefficients:
\begin{equation} \label{expert_sim}
\alpha_i = \frac{\exp\left( \text{sim}(\mathbf{d}^*, \mathbf{d}_i) / \tau \right)}{\sum_{j \in \mathcal{I}_{\text{top-}k}} \exp\left(\text{sim}(\mathbf{d}^*, \mathbf{d}_j) / \tau\right)}, \quad i \in \mathcal{I}_{\text{top-}k},
\end{equation}
where $\tau > 0$ is a temperature parameter. The semantic LoRA prior 
for task $\mathcal{T}^*$ is computed as a weighted sum:
\begin{equation} \label{prior}
\muv^* = \sum_{i \in \mathcal{I}_{\text{top-}k}} \alpha_i \cdot \muv_i,
\end{equation}

To better fit the prior distribution at the distributional level, we also consider estimating the variance for the unseen task as a function of expert parameters and their contribution weights:
\begin{equation} \label{prior}
{\sigmav^2}^* = \mathcal{F}\big(\muv^*, \{\boldsymbol{\mu}_i, \sigmav_i^2\}_{i=1}^K\big),
\end{equation}
where $\mathcal{F}(\cdot)$ denotes an aggregation function that combines expert-level uncertainties and deviations to capture the overall task variance. This flexible formulation enables the model to better reflect the statistical properties of new tasks, which is crucial for generative modeling. Leveraging semantic similarities, the model adapts more efficiently to novel tasks. The estimated mean and variance guide the generative process, resulting in more accurate and context-aware outputs, thereby improving generalization and robustness. To simplify, we will use 
$c$ to represent the semantic prior in the following descriptions.



\subsubsection{Conditional LoRA Parameter Generation}
We adopt a conditional variational autoencoder framework to generate target LoRA parameters based on task-aware prior information. Given a batch of training LoRA tensor $\Xv\in \mathbb{R}^{B \times m \times n}$, 
the encoder approximates the posterior distribution $q(\zv|\Xv, c)$ using a multi-layer perceptron (MLP) that takes the to be reconstructed $\Xv$ and the semantic prior $c$ as input. A latent code $z \sim q(\zv|\Xv, c)$ is sampled and passed to the decoder along with $c$ as condition to reconstruct the original input $\Xv$.
To model the task-conditional prior $p(\zv | c)$, we use a separate MLP, referred to as the prior mapper. This allows the model to flexibly represent a task-specific prior distribution based on domain-level statistics.

The model is trained to maximize the evidence lower bound
(ELBO), which consists of two terms: the reconstruction and the regularization term
:
\begin{equation} \label{KL}
\mathcal{L}_{\text{CVAE}} = \mathbb{E}_{q(\zv|\Xv, c)}\left[ \|\Xv - \hat{\Xv}\|^2 \right] + \lambda \cdot \mathrm{\textit{KL}}(q(\zv|\Xv, c) \| p(\zv| c)),
\end{equation}
where $\hat{\Xv}$ denotes the reconstructed LoRA parameters, $KL(\cdot \parallel \cdot)$ is the Kullback-Leibler divergence, and $\lambda$ controls the relative weight of the \textit{KL} term.
The first term encourages the decoder to reconstruct accurate LoRA parameters, 
while the second term regularizes the latent space to match the task-specific prior.
During inference, a sample $z$ is drawn from the prior distribution $ p(\zv| c)$, and the decoder generates the corresponding custom LoRA parameter.

\section{Experiments}

\subsection{Experimental Settings}
We conducted experiments on both image-text retrieval and cross-dataset generalization tasks.
Specifically, we use the widely adopted MS-COCO dataset \cite{lin2014microsoft}, a standard benchmark for image-caption retrieval, known for its diverse scenes and rich linguistic annotations.
To evaluate the model’s generalization ability across visual domains, we further include the OxfordPets dataset \cite{parkhi2012cats} and the Flowers102 dataset \cite{nilsback2008automated}, both of which are originally designed for fine-grained image classification.
Moreover, given the inherent ambiguity and limited informativeness of MS-COCO captions and the absence of captions in the other two datasets, we construct synthetic textual descriptions using  Qwen2-VL \cite{wang2024qwen2}. For MS-COCO, we use the original training split but regenerate captions for each image, effectively creating a new image–caption dataset, and then divide it into training, validation, and test sets. These datasets enable us to assess the model’s robustness both in in-domain retrieval and in generalizing to novel domains and compositional scenarios.
The evaluation metric used is Recall@K (R@K), which quantifies the proportion of correct matches appearing in the top-K retrieved candidates. We report R@1, R@5, and R@10 for both image-to-text and text-to-image retrieval scenarios.

\subsection{Implementation Details}
We adopt CLIP ViT-B/16 as our backbone, injecting rank-2 LoRA adapters into the $W_q$, $W_k$, and $W_v$ projection matrices of every Transformer block in the visual encoder. Training is carried out with the Adam optimizer. The CVAE’s encoder and prior network each consist of two-layer MLPs with ReLU activations, whereas the decoder is realized as a three-layer MLP with ReLU activations. We set the default values of $M$ and $k$ in the model to 100 and 4, respectively. All experiments were performed on a single NVIDIA A6000 GPU. 

\subsection{Comparative Methods}
To evaluate the effectiveness of the
proposed method, we compare it with following methods:
(1) \textbf{Zero-Shot CLIP}: The original CLIP model without any adaptation of LoRA modules.
(2) \textbf{Model Soups}: Consistent with \cite{wortsman2022model} all LoRA experts in $\mathcal{W}_{\text{expert}}$ are uniformly averaged without considering their relevance to the target task.
(3) \textbf{Top-\textit{k} LoRA Merging (Ours)}: The top-$k$ experts from the expert repository are selected based on the semantic similarity vector and with equal weight, assigning each a coefficient of $1/k$.
(4) \textbf{Top-\textit{k} LoRA Weighted (Ours)}: The top-$k$ experts are selected based on the semantic similarity vector, and their weights are computed by applying a softmax function over the similarity scores for adaptive merging. (5) \textbf{\texttt{SG-LoRA} (Ours)} Our proposed method, which generates
task-specific LoRAs based on semantic proximity. (6) \textbf{Oracle:} For each task, LoRA parameters are trained individually using the respective image-caption pair dataset.

\subsection{Main Results and Discussion}
\subsection{Image-Text Retrieval Within a Single Dataset}
We conducted image-text retrieval evaluations on both the MS-COCO and OxfordPets datasets, with results shown in Table.\ref{coco} -\ref{oxford}. Several key observations are as follows:
1). Compared to the Zero-shot CLIP baseline and consistent with previous findings \cite{qorbani2025semantic}, directly merging all experts in the expert repository leads to performance improvements.
2). Selecting semantically relevant experts—those most related to the current query task—from the repository can further enhance performance. However, naively treating all selected experts equally may result in degraded performance. For instance, in Table \ref{oxford}, uniform Top-\textit{k} LoRA Merging underperforms compared to Top-\textit{k} LoRA Weighted.
This may be because different experts contribute unevenly to the target task. Therefore, assigning equal weight ignores the varying degrees of semantic alignment, potentially amplifying noise from less relevant experts.
Conversely, incorporating semantic weighting coefficients allows for more effective integration of expert knowledge during fusion, thereby improving retrieval performance.
3). While merging-based approaches still fall short of the performance achieved by directly fine-tuning LoRA on the unseen task, \texttt{SG-LoRA} fully recovers the performance of oracle adapters.
Notably, \texttt{SG-LoRA} even outperforms task-specific fine-tuned LoRAs in certain cases—for example, R@1 and R@5 for bidirectional retrieval in Table.\ref{coco} and R@1 in for image-to-text retrieval in Table.\ref{oxford}.
This improvement gains from the efficient compression of expert LoRAs: our trained CVAE effectively integrates the target LoRA distribution using compact yet semantically rich task representations, enabling the generation of target-aligned LoRA parameters by modeling their distribution in the parameter space.
Moreover, the oracle LoRA fine-tuned on unseen tasks sometimes suffers from overfitting, especially when trained on a small set of image-caption pairs. Our \texttt{SG-LoRA} helps mitigate this performance drop, likely due to its ability to generalize without relying on target-task data.

\begin{table}[h]
\vspace{-2mm}
\centering
\small
\caption{Model Performance of image-text retrieval on MS-COCO. The best results are highlighted in bold, and the second-best results are underlined.}
\label{coco}
\begin{tabular}{l|ccc|ccc}
\toprule
\multirow{2}{*}{\textbf{Method}} & \multicolumn{3}{c|}{\textbf{I2T Metrics}} & \multicolumn{3}{c}{\textbf{T2I Metrics}} \\
\cmidrule(lr){2-4} \cmidrule(lr){5-7}
 & \textbf{R@1} & \textbf{R@5} & \textbf{R@10} & \textbf{R@1} & \textbf{R@5} & \textbf{R@10} \\
\midrule
Zero-shot CLIP & 66.43 & 84.31 & 89.14 & 41.66 & 64.63 & 73.01 \\
Oracle & 72.45 & 88.91 & 93.41 & 53.10 & 76.47 & 83.97 \\
\midrule
Model Soups & 69.37 & 85.96 & 90.95 & 47.38 & 69.54 & 77.97 \\
Top-\textit{k} LoRA Merging & 70.70 & 86.57 & 91.09 & 48.64 & 70.51 & 78.79 \\
Top-\textit{k} LoRA Weighted& \underline{71.55} & \underline{87.54} & \underline{91.69} & \underline{49.85} & \underline{71.79} & \underline{79.66} \\
\texttt{SG-LoRA} & \textbf{74.31} & \textbf{88.78} & \textbf{92.50} & \textbf{54.42} & \textbf{75.45} & \textbf{82.18} \\
\midrule
\end{tabular}
\end{table}

\begin{table}[h]
\centering
\small
\vspace{-5mm}
\caption{
Model Performance of image-text retrieval on OxfordPets. The best results are highlighted in bold, and the second-best results are underlined.}
\label{oxford}
\begin{tabular}{l|ccc|ccc}
\toprule
\multirow{2}{*}{\textbf{Method}} & \multicolumn{3}{c|}{\textbf{I2T Metrics}} & \multicolumn{3}{c}{\textbf{T2I Metrics}} \\
\cmidrule(lr){2-4} \cmidrule(lr){5-7}
 & \textbf{R@1} & \textbf{R@5} & \textbf{R@10} & \textbf{R@1} & \textbf{R@5} & \textbf{R@10} \\
\midrule
Zero-shot CLIP & 40.45 & 66.27 & 77.53 & 26.03 & 50.66 & 62.98 \\
Oracle & {55.84} & {81.84} & {89.13} & {40.99} & {70.41} & {80.39} \\
\midrule
Model Soups & 52.54 & 77.80 & 85.59 & 33.51 & 61.77 & 72.93 \\
Top-\textit{k} LoRA Merging & 52.59 & 78.52 & 86.09 & 34.05 & 62.70 & 73.93 \\
Top-\textit{k} LoRA Weighted & \underline{53.96} & \underline{79.41} & \underline{86.53} & \underline{35.42} & \underline{64.08} & \underline{74.99} \\
\texttt{SG-LoRA} & \textbf{57.15} & \textbf{80.40} &  \textbf{88.04} &  \textbf{37.62} & \textbf {67.16} & \textbf{77.44} \\
\midrule
\end{tabular}
\end{table}

\subsection{Cross-Dataset Image-Text Retrieval}
Moreover, given the flexibility of \texttt{SG-LoRA}, we conducted a more challenging cross-dataset evaluation. As shown in Table.\ref{coco-pets}-\ref{coco-flowers}, \texttt{SG-LoRA} consistently outperforms merging-based approaches in these settings.
Interestingly, we also observed that models trained on MS-COCO were able to generate LoRA parameters that, in some cases, outperformed those trained directly on the OxfordPets dataset. This may be attributed to the richer expert knowledge available in MS-COCO, which, due to its broader data diversity, enables more extensive exploration of the parameter space during generation—something not achievable when generating within the narrower scope of OxfordPets.
Another possible reason is that the uniform LoRA training configuration used across datasets (as described in Section \ref{ss:loratraining}) may not be optimal for OxfordPets.
Conversely, comparing Table.\ref{pets-coco} with Table.\ref{coco}, we find that when the generation model is trained on OxfordPets and applied to MS-COCO, its performance is generally worse than that of models trained on MS-COCO. This is likely due to the limited expert diversity in OxfordPets, which focuses primarily on fine-grained distinctions within just two broad categories—cats and dogs—thus making it less effective at generating LoRAs for more diverse tasks found in MS-COCO.

\subsection{Generalization in Standard Image-Text Retrieval}
\begin{figure}[!t]
\centering
\includegraphics[width=0.80\linewidth]{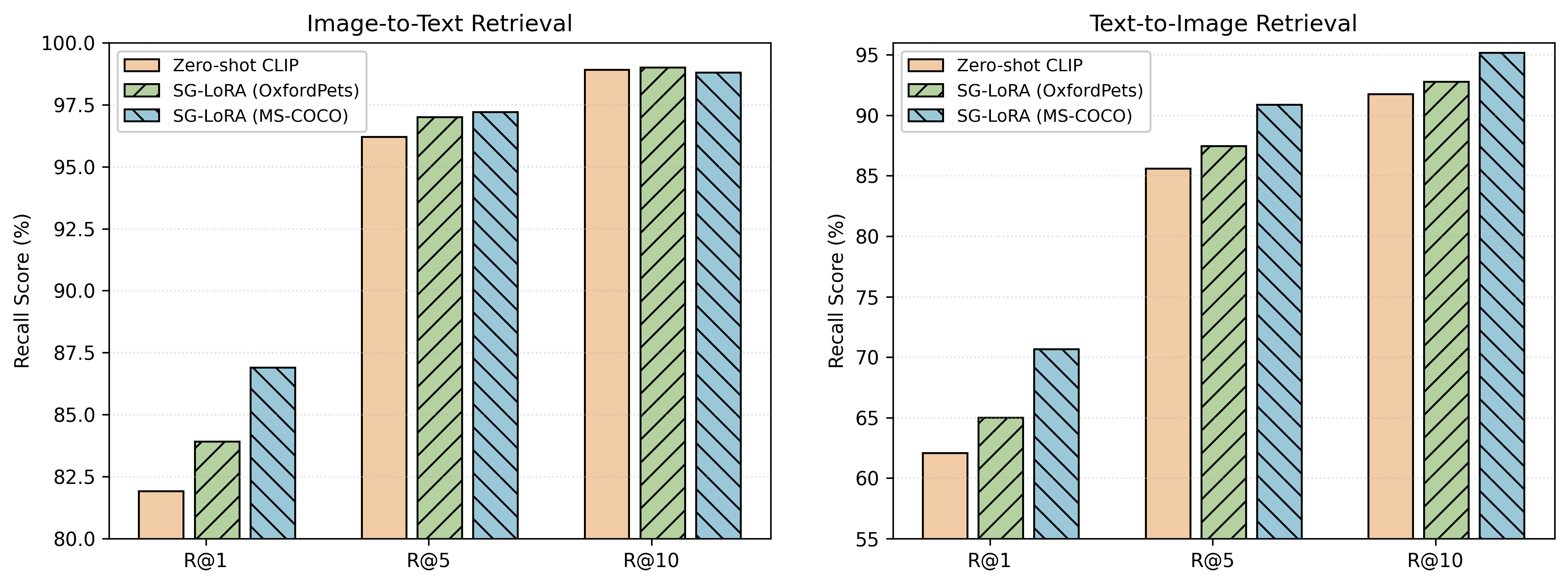}
\caption{\small{Simulation of task-agnostic evaluation on a general image-text retrieval dataset.
We train models on MS-COCO and OxfordPets respectively, and evaluate them on the Flickr30K test set. To mimic the scenario where the task description is not explicitly given, we randomly sample one caption per image from Flickr30K and use the CLIP textual encoder to obtain embeddings for all captions, which are then averaged to construct the task description.}}
\label{flickr}
\end{figure}

Considering another complex scenario where the test task may contain image-caption pairs of multiple categories, we evaluated the retrieval performance on the test set of the Flickr30K dataset \cite{plummer2015flickr30k}. Since this dataset has no clear category distinction during retrieval, we randomly selected one caption from the test set, fed it into the CLIP textual encoder to obtain the textual embedding, and then calculated the mean value as the task description for retrieval.
The experimental results are shown in Figure.\ref{flickr}, where \texttt{SG-LoRA} outperforms Zero-shot CLIP. Additionally, \texttt{SG-LoRA} trained on MS-COCO achieves better results than that trained on OxfordPets. This is because MS-COCO provides more comprehensive expert knowledge, covering a wider range of categories, while OxfordPets primarily focuses on fine-grained knowledge of cat and dog categories. This also indicates that when semantic guidance is more powerful and comprehensive, or more relevant to the downstream task, the generated LoRA parameters are also superior.

\begin{table}[h]
\centering
\small
\caption{Cross-dataset evaluation with the model trained on MS-COCO and tested on OxfordPets. The best results are highlighted in bold, and the second-best results are underlined.}
\label{coco-pets}
\begin{tabular}{l|ccc|ccc}
\toprule
\multirow{2}{*}{\textbf{Method}} & \multicolumn{3}{c|}{\textbf{I2T Metrics}} & \multicolumn{3}{c}{\textbf{T2I Metrics}} \\
\cmidrule(lr){2-4} \cmidrule(lr){5-7}
 & \textbf{R@1} & \textbf{R@5} & \textbf{R@10} & \textbf{R@1} & \textbf{R@5} & \textbf{R@10} \\
\midrule
Zero-shot CLIP & 40.45 & 66.27 & 77.53 & 26.03 & 50.66 & 62.98 \\
Oracle & {55.84} & {81.84} & {89.13} & {40.99} & {70.41} & {80.39} \\
\midrule
Model Soups & 44.67 & 70.91 & 80.78 & 30.45 & 56.77 & 68.52 \\
Top-\textit{k} LoRA Merging & 45.96 & 71.83 & 81.42 & 30.88 & 57.32 & 69.08 \\
Top-\textit{k} LoRA Weighted & \underline{48.13} & \underline{73.43} & \underline{82.73} & \underline{33.34} & \underline{59.53} & \underline{70.89} \\
\texttt{SG-LoRA} & \textbf{55.41} & \textbf{80.73} & \textbf{87.33} & \textbf{38.84} & \textbf{66.77} & \textbf{76.69} \\
\midrule
\end{tabular}
\end{table}

\begin{table}[h]
\centering
\small
\caption{Cross-dataset evaluation with the model trained on OxfordPets and tested on MS-COCO. The best results are highlighted in bold, and the second-best results are underlined.
}
\label{pets-coco}
\begin{tabular}{l|ccc|ccc}
\toprule
\multirow{2}{*}{\textbf{Method}} & \multicolumn{3}{c|}{\textbf{I2T Metrics}} & \multicolumn{3}{c}{\textbf{T2I Metrics}} \\
\cmidrule(lr){2-4} \cmidrule(lr){5-7}
 & \textbf{R@1} & \textbf{R@5} & \textbf{R@10} & \textbf{R@1} & \textbf{R@5} & \textbf{R@10} \\
\midrule
Zero-shot CLIP & 66.43 & 84.31 & 89.14 & 41.66 & 64.63 & 73.01 \\
Oracle & 72.45 & 88.91 & 93.41 & 53.10 & 76.47 & 83.97 \\
\midrule
Model Soups & 68.58 & 85.67 & 90.62 & 44.09 & 66.55 & 75.08 \\
Top-\textit{k} LoRA Merging & {68.74} & \underline{85.83} & {90.63} & {44.19} & {66.58} & \underline{75.31} \\
Top-\textit{k} LoRA Weighted & \underline{68.75} & 85.77 & \underline{90.67} & \underline{44.60} & \underline{66.76} & 75.25 \\
\texttt{SG-LoRA} & 
\textbf{70.81} & \textbf{86.83} & \textbf{91.41} & \textbf{46.50} & \textbf{68.73} & \textbf{77.19} \\
\midrule
\end{tabular}
\end{table}

\begin{table}[h]
\centering
\small
\vspace{-2mm}
\caption{Cross-dataset evaluation with the model trained on MS-COCO and tested on Flowers102. The best results are highlighted in bold, and the second-best results are underlined.}
\label{coco-flowers}
\begin{tabular}
{l|ccc|ccc}
\toprule
\multirow{2}{*}{\textbf{Method}} & \multicolumn{3}{c|}{\textbf{I2T Metrics}} & \multicolumn{3}{c}{\textbf{T2I Metrics}} \\
\cmidrule(lr){2-4} \cmidrule(lr){5-7}
 & \textbf{R@1} & \textbf{R@5} & \textbf{R@10} & \textbf{R@1} & \textbf{R@5} & \textbf{R@10} \\
\midrule
Zero-shot CLIP & 22.30 & 53.57 & 69.41 & 16.33 & 46.79 & 68.49 \\
Oracle & {26.23} & {59.17} & {77.55} & {21.57} & {57.56} & {80.13} \\
\midrule
Model Soups & 23.50 & \underline{54.84} & \underline{73.85} & 17.23 & \underline{50.96} & 73.48 \\
Top-\textit{k} LoRA Merging & \underline{24.21} & 54.26 & {72.83} & \underline{17.76} & {50.93} & \underline{73.83} \\
Top-\textit{k} LoRA Weighted & {23.98} & {52.83} & 71.91 & {17.63} & 50.69 & 73.74 \\
\texttt{SG-LoRA} & \textbf{26.83} & \textbf{56.63} & \textbf{74.16} & \textbf{20.52} & \textbf{53.71} & \textbf{76.69} \\
\midrule
\end{tabular}
\end{table}

\subsection{Ablation Stduy}






\paragraph{Impact of expert repository configuration.}
In Table \ref{coco-pets}, we observed that the \textit{Cat} expert from the MS-COCO dataset were frequently selected as semantic conditions. To further evaluate the impact of semantically salient experts, we assessed how the inclusion of the \textit{Cat} expert from MS-COCO influences retrieval performance on two unseen cat classes in the OxfordPets dataset.
As shown in Table.
\ref{cat expert}, incorporating the \textit{Cat} expert into the expert repository improves performance in most cases, particularly for text-to-image retrieval. This highlights the effectiveness of semantically guided expert selection. This finding also demonstrates the flexibility of our method in constructing task-adaptive expert repositories—particularly when a richer pool of LoRA resources is available.


\begin{table}[h]
\centering
\small
\caption{Ablation study on expert repository strategy for cross-dataset evaluation.
We assess the impact of the \textit{Cat} expert from MS-COCO dataset on the retrieval performance in two unseen cat tasks from OxfordPets dataset.}
\label{cat expert}
\begin{tabular}{l|ccc|ccc}
\toprule
\multirow{2}{*}{\centering\textbf{Expert}} & \multicolumn{3}{c|}{\textbf{I2T Metrics}} & \multicolumn{3}{c}{\textbf{T2I Metrics}} \\
\cmidrule(lr){2-4} \cmidrule(lr){5-7}
\centering\textbf{strategy} & \textbf{R@1} & \textbf{R@5} & \textbf{R@10} & \textbf{R@1} & \textbf{R@5} & \textbf{R@10} \\
\midrule
\multicolumn{7}{c}{\textcolor{gray}{\textit{Egyptian Mau}}} \\
\midrule
w/o \textit{Cat} expert & {36.08} & {62.89} & {71.13} & {15.21} & {31.70} & {44.07} \\
w \textit{Cat} expert & 37.11 & 63.92 & 72.16 & 15.21 & 35.05 & 46.91 \\
\midrule
\multicolumn{7}{c}{\textcolor{gray}{\textit{Persian}}} \\
\midrule
w/o \textit{Cat} expert & {44.00} & {80.00} & {87.00} & {34.00} & {62.25} & {72.75} \\
w \textit{Cat} expert
& 47.00 & 79.65 & 86.00 & 36.75 & 64.25 & 73.75 \\
\midrule
\end{tabular}
\end{table}

\begin{table}[htbp]
\centering
\small
\caption{Top-4 expert for \textit{Yorkshire Terrier} category selected by semantic similarity under mixed-source experts configurations}
\label{tab:yorkshire_terrier}
\begin{tabular}{lcl}
\toprule
\textbf{Expert Task} & \textbf{Source Dataset} & \textbf{Contribution} \\
\midrule
\textit{Scottish Terrier} & OxfordPets & 0.9221 \\
\textit{Dog}             & MS-COCO    & 0.0692 \\
\textit{Cat}             & MS-COCO    & 0.0082 \\
\textit{American Bulldog} & OxfordPets & 0.0005 \\
\bottomrule
\end{tabular}
\end{table}

\begin{figure}[!t]
\centering
\includegraphics[width=0.80\linewidth]{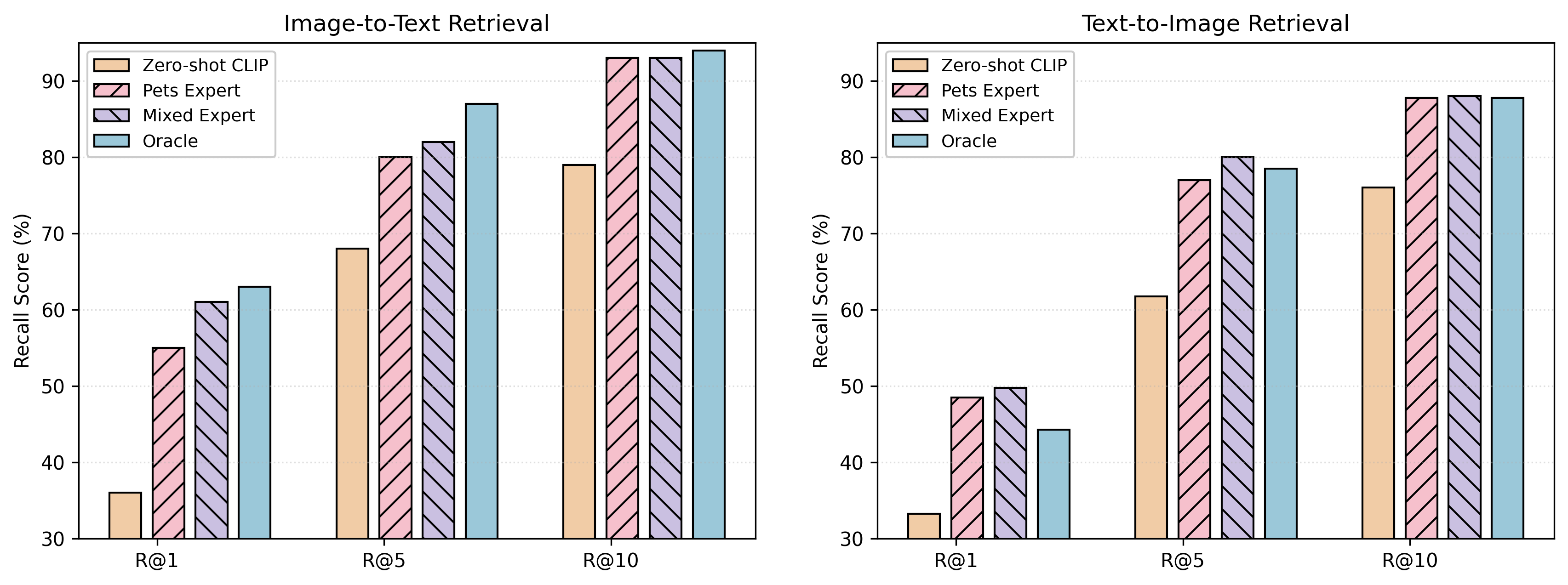}
\caption{\small{Comparison on expert repository configuration: single-source experts(OxfordPets) vs. mixed-source experts.}}
\label{single-mix expert}
\end{figure}

Given that our model supports open-world expert repository construction, we further conducted a case study in which experts were selected from the OxfordPets and MS-COCO and Flowers102 datasets to form a mixed-source expert repository. We also combined training data from both datasets to train the \texttt{SG-LoRA} model accordingly.
Figure.\ref{single-mix expert} presents a comparison between single-source and mixed-source expert configurations. Additionally, we present the top-4 experts selected by \texttt{SG-LoRA} under the mixed-source setup for the unseen \textit{Yorkshire Terrier} category, along with their corresponding weights. As shown, the mixed-source experts yield slightly better performance for the \textit{Yorkshire Terrier} category. The performance even surpasses that of the oracle LoRA model in text-to-image retrieval.
These demonstrates the potential of our method in more realistic, real-time application scenarios, where expert repositories are constructed dynamically from heterogeneous data sources.

\begin{figure}[!t]
\centering
\includegraphics[width=1.0\linewidth]{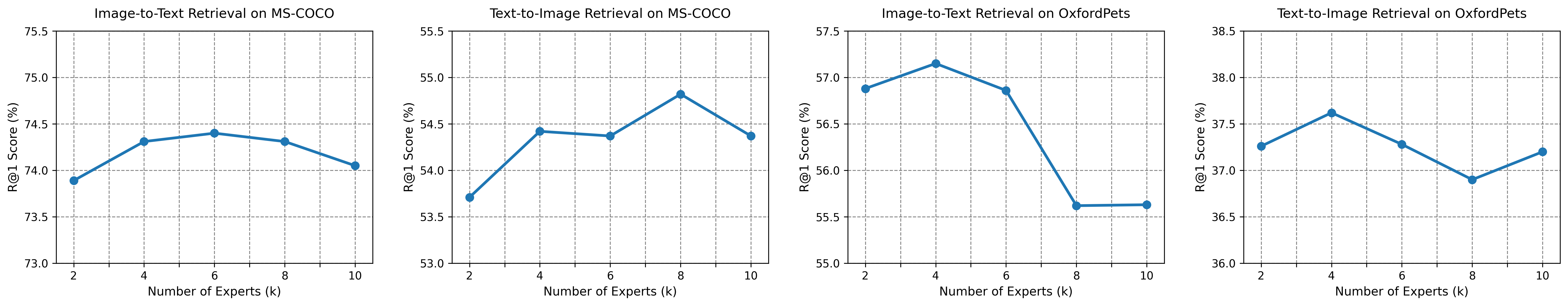}
\caption{\small{Ablaton study on the number of expert.}}
\label{expert number}
\end{figure}

\paragraph{Impact of the number of expert.}
We conduct ablation studies on the number of experts $k$ used in semantic prior construction. As shown in Figure.\ref{expert number}, using too few experts leads to insufficient knowledge for generalizing to unknown tasks, while increasing $k$ incorporates more semantic information but may also introduce irrelevant context. Overall, $k=4$ shows a good balance between expert diversity and relevance, delivering good performance on both datasets.

\section{Conclusion}

In this work, we introduce a novel and challenging task setting—Zero-Shot Open-world Adaptation (ZSOA)—which requires models to generalize across isomorphic yet semantically diverse tasks in open-world scenarios. To tackle this, we propose a flexible and efficient approach that dynamically generates task-specific LoRA parameters guided by available task semantics. By identifying the most relevant expert knowledge based on semantic similarity and leveraging weighted combinations in a conditional generative framework, our \texttt{SG-LoRA} models the distribution of target parameters without the need for traditional fine-tuning. The inherent stochasticity of our generation process further introduces diversity, enhancing adaptability to previously unseen tasks.
\texttt{SG-LoRA} is scalable and naturally privacy-preserving, making it well-suited for deployment in sensitive and dynamic real-world environments.

\bibliography{iclr2025_conference}
\bibliographystyle{iclr2025_conference}

\clearpage

\appendix
\section{Appendix}
\subsection{Experimental Details}
\begin{figure}[!ht]
\centering
\includegraphics[width=0.85\linewidth]{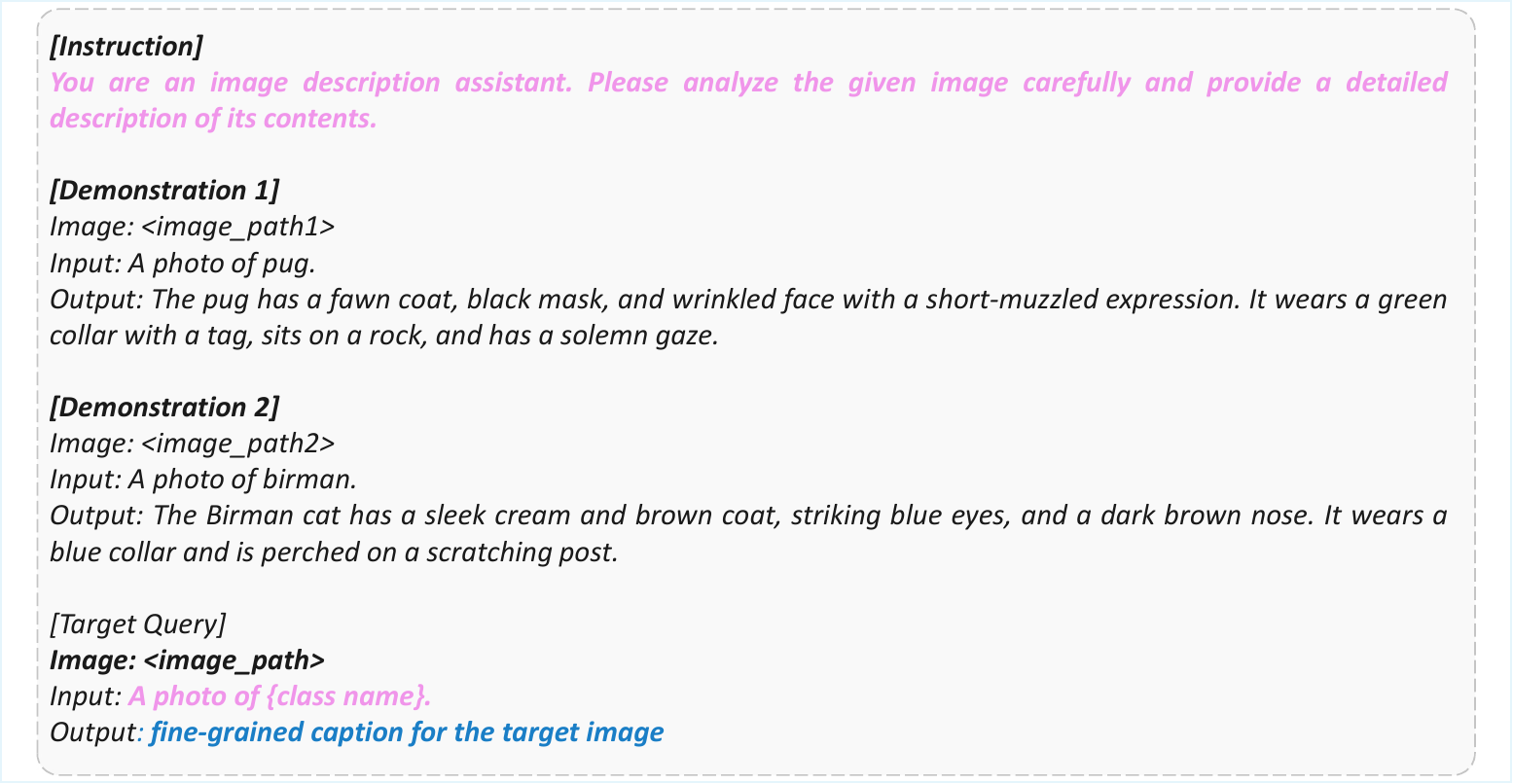}
\caption{\small{Examples of caption generation using Qwen2-VL.}}
\label{captions}
\end{figure}

In this work, we generate fine-grained captions for the image-text retrieval task using Qwen2-VL. As illustrated in Figure.\ref{captions}, our in-context learning approach provides the Multimodal Large Language Model (MLLM) with a small number of demonstration examples, enabling it to generate detailed and accurate captions for target images. Using this method, we have produced diverse and highly relevant captions for the entire OxfordPets dataset and the Flowers102 dataset, a subset of the MS-COCO dataset. These fine-grained descriptions serve as a valuable resource for downstream tasks such as fine-grained image-text retrieval and facilitate further research in ZSOA.

\begin{figure}[!ht]
\centering
\includegraphics[width=1\linewidth]{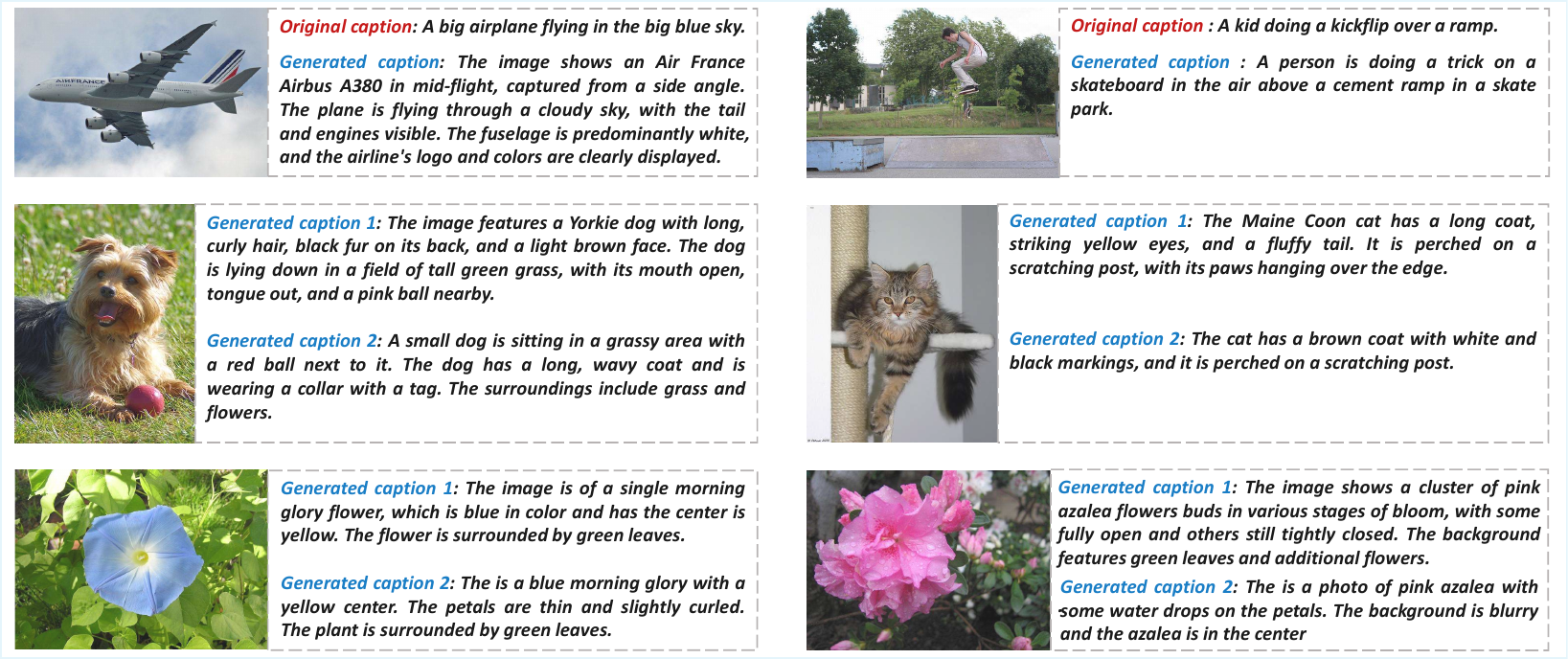}
\caption{\small{Illustration of generated captions: results on the MS-COCO (top), OxfordPets (middle), and Flowers102 (bottom) datasets.}}
\label{captions_show}
\end{figure}

As shown in Figure.\ref{captions_show}, we present examples of image-caption pairs from three datasets. Compared to the original captions in the COCO dataset (labeled as \textit{Original Caption} in the top), our generated captions more accurately reflect the content of the corresponding images.

\subsection{Additional Results on Classification Task}
\begin{figure}[!ht]
\centering
\includegraphics[width=0.9\linewidth]{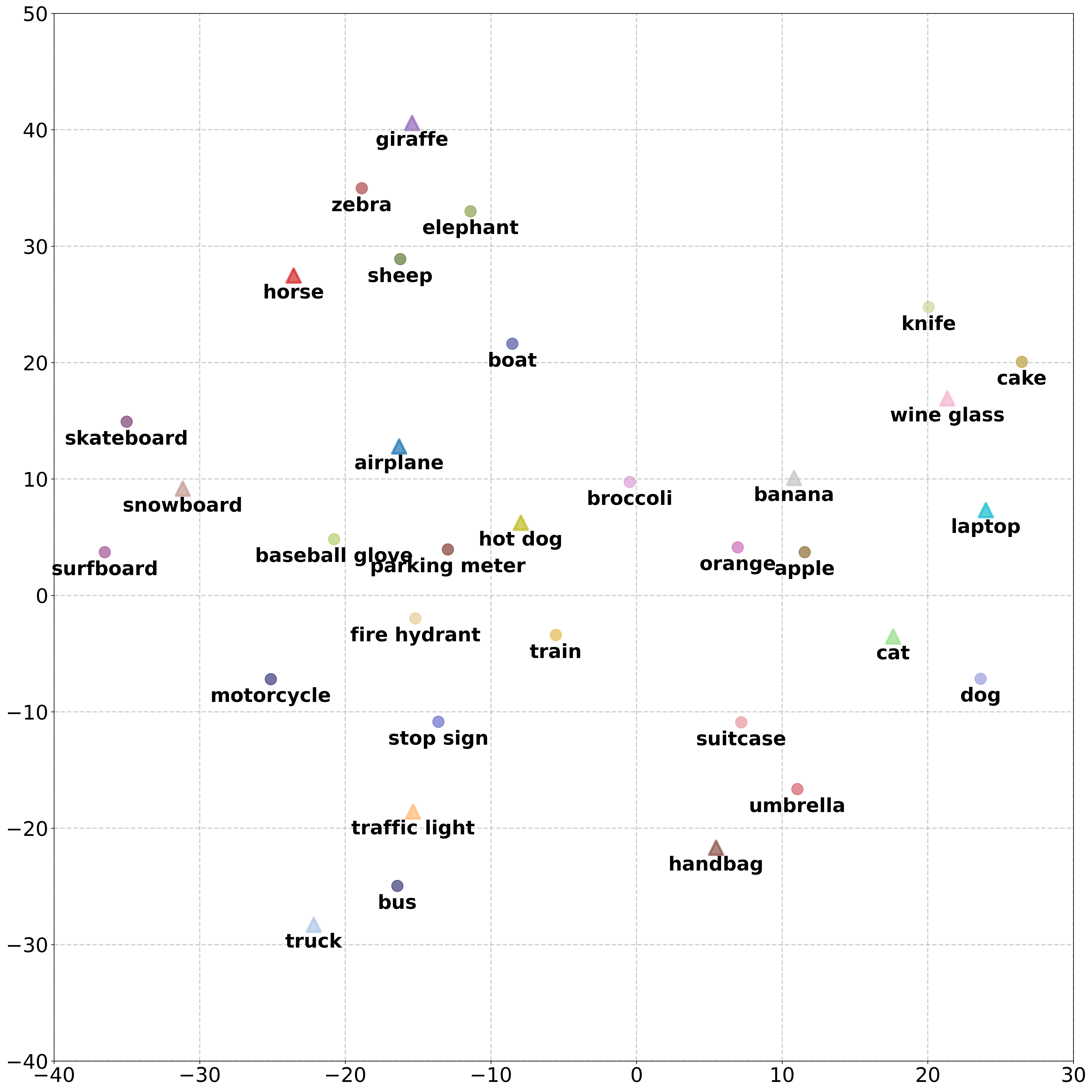}
\caption{\small{t-SNE visualization of the averaged LoRA parameters on MS-COCO dataset for image-text retrieval task. Triangular markers indicate expert LoRAs. Semantically similar LoRA parameters tend to cluster closely together.}}
\label{t-sne_coco}
\end{figure}

\begin{figure}[!ht]
\centering
\includegraphics[width=0.9\linewidth]{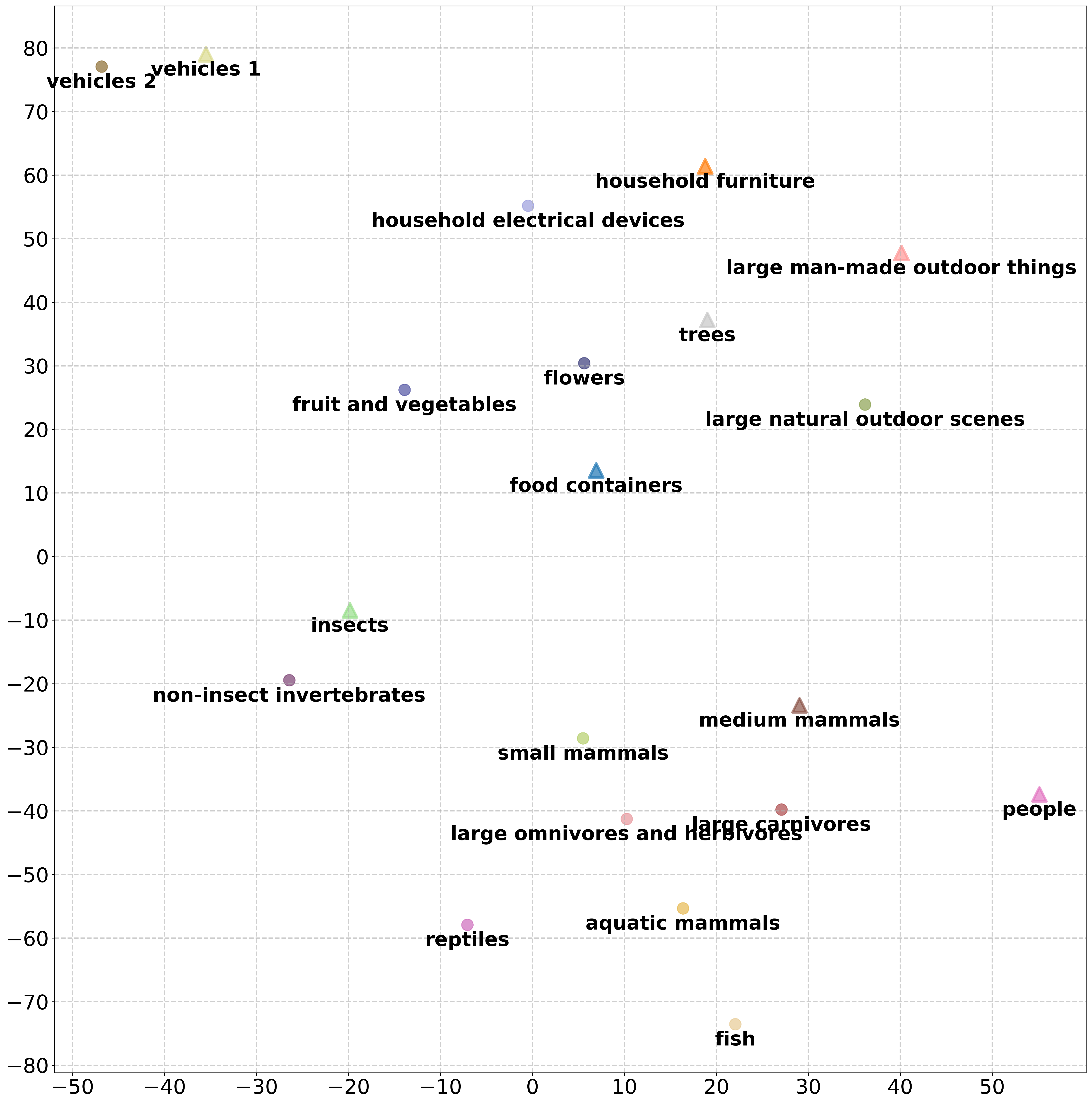}
\caption{\small{t-SNE visualization of the averaged LoRA parameters on CIFAR-100 for classification task. Triangular markers indicate expert LoRAs. Although semantically similar LoRA parameters tend to cluster closely together, the distribution appears sparser compared to the MS-COCO dataset due to stronger task independence.}}
\label{t-sne_cifar}
\end{figure}

In addition to the image-text retrieval task, we also explored the performance of \texttt{SG-LoRA} on classification tasks. Specifically, we selected 20 superclasses from CIFAR-100 as 20 distinct tasks, with each task corresponding to a 5-class classification. We selected 8 of these tasks as expert LoRAs and performed inference on 6 unseen tasks, with the results presented in Table.\ref{CIFAR-100}. We observed that, compared to Zero-shot CLIP, both Model Soups and the selection of expert parameters to construct LoRA improved classification performance on unseen tasks. Furthermore, \texttt{SG-LoRA} achieved best performance, indicating that our method is also applicable to classification tasks. However, we found that \texttt{SG-LoRA} still has a significant performance gap compared to fine-tuning LoRA directly on unseen tasks using training data. We hypothesize that this is because, for classification tasks, the independence (or orthogonality, where decision boundaries between tasks may be orthogonal) between tasks is much more pronounced. In contrast, for image-text retrieval tasks, which operate at a finer granularity, there exist stronger inter-task correlations—this inherent property facilitates feasible LoRA parameter transfer. As shown in Figure.\ref{t-sne_cifar}, we visualized the average LoRA parameters for the 20 tasks and found that, although semantically similar tasks are closer in parameter space, their distances remain relatively sparse compared to image-text retrieval tasks in Figure.\ref{t-sne_coco}.We plan to conduct further exploration on this issue in the future.

Additionally, we conducted an ablation study on textual description for the classification task, as shown in Table.\ref{tab:cifar100_cap}. It can be observed that, for the current task setting, more detailed textual descriptions, which account for the specific categories within each task, can better capture the semantic relationships between tasks, thus leading to improved performance.

\begin{table}[h]
\centering
\caption{Model Performance of image classification on CIFAR-100 superclass.}
\label{CIFAR-100}
\begin{tabular}{l|c}
\toprule
\centering
\textbf{Method} & 
 \textbf{Accuracy}  \\
\midrule
Zero-shot CLIP & 72.30  \\
Oracle & 91.43  \\
\midrule
Model Soups & 75.63  \\
Top-\textit{k} LoRA Merging & 72.60  \\
Top-\textit{k} LoRA Weighted & 72.70 \\
\texttt{SG-LoRA} & 77.50 \\
\midrule
\end{tabular}
\end{table}

\begin{table}[htbp]
\centering
\caption{
Ablation study
of textual description on CIFAR-100 superclass Classification}
\label{tab:cifar100_cap}
\begin{tabular}{@{}p{8cm}c@{}}
\toprule
\multicolumn{1}{c}{\textbf{Description}} & \textbf{Accuracy (\%)} \\ 
\midrule
A photo of a \textit{<superclass name>} & 75.77 \\
\midrule
\makecell[l]{This is a classification task for recognizing \textit{<superclass name>}, \\ 
which includes \textit{class\_1, ..., class\_5}} & \multirow{2}{*}{77.50} \\
\bottomrule
\end{tabular}
\end{table}

\subsection{More Ablations and Analysis}

\begin{table}[htbp]
\centering
\caption{Ablation study on modalities of semantic prior condition}
\label{tab:vis_txt}
\begin{tabular}{lccccc}
\toprule
\multirow{2}{*}{Condition} & 
\multicolumn{2}{c}{Metrics} & & \multirow{2}{*}{Dataset} \\
\cmidrule{2-3}
& I2T R@1 & T2I R@1 & & \\
\midrule
Visual & 73.16 & 52.70 & &\multirow{2}{*}{MS-COCO} \\
\cellcolor{gray!10}{Textual} & \cellcolor{gray!10}74.31 & \cellcolor{gray!10}54.42 & & \\
\addlinespace[2pt]
Visual & 86.30 & 70.12 & & \multirow{2}{*}{Flickr30K} \\
\cellcolor{gray!10}Textual & \cellcolor{gray!10}86.90 & \cellcolor{gray!10}70.66 & & \\
\midrule
\multirow{2}{*}{Condition} & 
\multicolumn{2}{c}{Metrics} & & \multirow{2}{*}{Dataset} \\
\cmidrule{2-3}
& \multicolumn{2}{c}{Accuracy} & & & \\
\midrule
Visual & \multicolumn{2}{c}{73.83} & & \multirow{2}{*}{CIFAR-100} \\
\cellcolor{gray!10}Textual & \multicolumn{2}{c} {\cellcolor{gray!10}77.50} & & & \\
\bottomrule
\end{tabular}
\end{table}

\paragraph{Ablation on modalities of semantic prior condition.}
The construction of semantic priors serves as the foundation for our \texttt{SG-LoRA}. In Table.\ref{tab:vis_txt}, we compare the performance results of prior conditions across different modalities. The visual condition is obtained by averaging the visual embeddings of training set images within each task dataset using a frozen CLIP visual encoder. Experimental results show that the textual condition better captures the semantic relationships between tasks. 
This could be attributed to two factors. Firstly, the high degree of condensation in textual semantics might play a role. Secondly, the disparities between training and test set images (or the presence of noisy images) within a task could result in inaccuracies in the visual prior condition.

\paragraph{Impact of expert repository configuration.}
Consistent with Table \ref{cat expert}, we further evaluated the impact of incorporating the \textit{Dog} expert from the MS-COCO dataset on retrieval performance for two unseen dog tasks in the OxfordPets dataset. As demonstrated in Table \ref{dog expert}, including the \textit{Dog} expert in the expert repository consistently improves the performance.

\begin{figure}[!ht]
\centering
\includegraphics[width=0.85\linewidth]{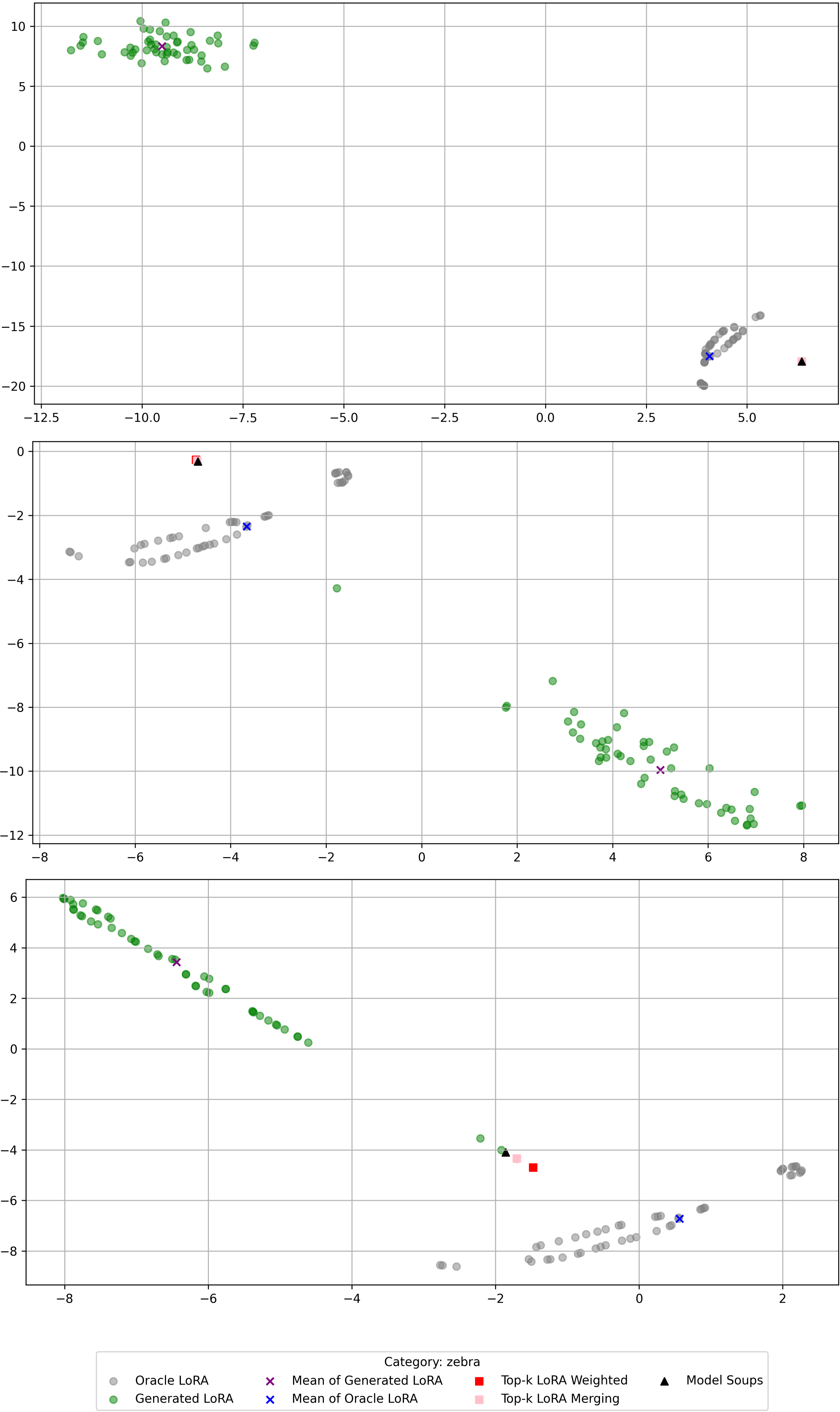}
\caption{\small{t-SNE visualization of LoRA parameters using different comparative methods, tested on the unseen '\textit{Zebra}' category at different training stages. For \texttt{SG-LoRA} and Oracle LoRA, we randomly sampled 50 samples each. The subfigures from top to bottom represent increasing CVAE training epochs.}}
\label{t-sne_zebra}
\end{figure}

\paragraph{Visulization of LoRA parameters.}
To further investigate the parameter diversity of \texttt{SG-LoRA}, we conducted evaluations on the unseen '\textit{Zebra}' task from the MS-COCO dataset at different training stages and visualized the generated LoRAs using t-SNE. As shown in Figure~\ref{t-sne_zebra}, we observe that the distribution of LoRAs generated by \texttt{SG-LoRA} gradually aligns with that of Oracle LoRAs (directly trained in image-caption pairs), while still preserving diversity rather than extensively overlapping with the Oracle. This indicate that, by injected stochasticity, our method effectively explores the high-performance LoRAs in the parameter space.

Additionally, by comparing the bottom subfigure, we can see that compared to Model Soup, which treats all experts equally, both  Tok-\textit{k} LoRA Merging and Tok-\textit{k} LoRA Weighted leverage semantic guidance to customize LoRA parameters better suired for the current unseen task. Thus, in parameter space, they lie closer to the mean value of the Oracle LoRA.

\begin{table}[h]
\centering
\small
\caption{Ablation study on expert repository strategy for cross-dataset evaluation.
We assess the impact of the \textit{Dog} expert from MS-COCO dataset on the retrieval performance in two unseen dog tasks from OxfordPets dataset.}
\label{dog expert}
\begin{tabular}{l|ccc|ccc}
\toprule
\multirow{2}{*}{\centering\textbf{Expert}} & \multicolumn{3}{c|}{\textbf{I2T Metrics}} & \multicolumn{3}{c}{\textbf{T2I Metrics}} \\
\cmidrule(lr){2-4} \cmidrule(lr){5-7}
\centering\textbf{strategy} & \textbf{R@1} & \textbf{R@5} & \textbf{R@10} & \textbf{R@1} & \textbf{R@5} & \textbf{R@10} \\
\midrule
\multicolumn{7}{c}{\textcolor{gray}{\textit{Pug}}} \\
\midrule
w/o \textit{Dog} expert & {45.00} & {66.00} & {78.00} & {32.00} & {51.00} & {61.25} \\
w \textit{Dog} expert & 46.00 & 67.00 & 78.00 & 32.00 & 51.75 & 62.75 \\
\midrule
\multicolumn{7}{c}{\textcolor{gray}{\textit{Chihuahua}}} \\
\midrule
w/o \textit{Dog} expert & {64.00} & {89.00} & {93.00} & {52.50} & {81.00} & {88.25} \\
w \textit{Dog} expert
& 66.00 & 90.00 & 95.00 & 55.25 & 81.25 & 89.75 \\
\midrule
\end{tabular}
\end{table}

\end{document}

%% file: math_commands.tex
\usepackage{amsmath,amsfonts,bm}

\def\eqref#1{equation~\ref{#1}}

\def\1{\bm{1}}

\DeclareMathAlphabet{\mathsfit}{\encodingdefault}{\sfdefault}{m}{sl}
\SetMathAlphabet{\mathsfit}{bold}{\encodingdefault}{\sfdefault}{bx}{n}

